\documentclass{article}

\PassOptionsToPackage{numbers, compress}{natbib}
\usepackage{cite}

\usepackage[final]{neurips_2023}

\usepackage[utf8]{inputenc} %
\usepackage[T1]{fontenc}    %
\usepackage[hidelinks,colorlinks=true,citecolor=darkgreen]{hyperref}       %
\usepackage{url}            %
\usepackage{booktabs}       %
\usepackage{amsfonts}       %
\usepackage{nicefrac}       %
\usepackage{microtype}      %

\usepackage{times}
\usepackage{epsfig}
\usepackage{graphicx}
\usepackage{amsmath}
\usepackage{amssymb}
\usepackage{listings}
\usepackage[dvipsnames]{xcolor}
\usepackage{booktabs}

\definecolor{codegreen}{rgb}{0,0.6,0}
\definecolor{codegray}{rgb}{0.5,0.5,0.5}
\definecolor{codepurple}{rgb}{0.58,0,0.82}
\definecolor{backcolour}{rgb}{0.95,0.95,0.92}

\lstdefinestyle{mystyle}{
  backgroundcolor=\color{backcolour},
  commentstyle=\color{codegreen},
  keywordstyle=\color{magenta},
  numberstyle=\tiny\color{codegray},
  stringstyle=\color{codepurple},
  basicstyle=\ttfamily\footnotesize,
  breakatwhitespace=false,         
  breaklines=true,                 
  captionpos=b,                    
  keepspaces=true,                 
  numbers=left,                    
  numbersep=5pt,                  
  showspaces=false,                
  showstringspaces=false,
  showtabs=false,                  
  tabsize=2
}

\usepackage[capitalize]{cleveref}
\crefname{section}{Sec.}{Secs.}
\Crefname{section}{Section}{Sections}
\Crefname{table}{Table}{Tables}
\crefname{table}{Tab.}{Tabs.}

\definecolor{ourPurple}{HTML}{9673A6}
\definecolor{ourOrange}{HTML}{D79B00}
\definecolor{ourGreen}{HTML}{82B366}
\definecolor{ourRed}{HTML}{B85450}
\definecolor{personColor}{HTML}{0000FF}
\definecolor{bgColor}{HTML}{bed4f3}

\title{Do Multilingual LLMs have specialized language heads?}

\author{%
  Muhammad Naufil \\
  Department of Computer Science\\
  Saarland University\\
  Saarbrucken, 66125 \\
  \texttt{muna00001@stud.uni-saarland.de} \\
}

\DeclareMathOperator*{\MHAtt}{MHAtt}

\newcommand{\qed}{\nobreak \ifvmode \relax \else
      \ifdim\lastskip<1.5em \hskip-\lastskip
      \hskip1.5em plus0em minus0.5em \fi \nobreak
      \vrule height0.75em width0.5em depth0.25em\fi}

\makeatletter
\def\thickhline{%
  \noalign{\ifnum0=`}\fi\hrule \@height \thickarrayrulewidth \futurelet
   \reserved@a\@xthickhline}
\def\@xthickhline{\ifx\reserved@a\thickhline
               \vskip\doublerulesep
               \vskip-\thickarrayrulewidth
             \fi
      \ifnum0=`{\fi}}
\makeatother

\makeatletter
\def\thickhline{%
  \noalign{\ifnum0=`}\fi\hrule \@height \thickarrayrulewidth \futurelet
   \reserved@a\@xthickhline}
\def\@xthickhline{\ifx\reserved@a\thickhline
               \vskip\doublerulesep
               \vskip-\thickarrayrulewidth
             \fi
      \ifnum0=`{\fi}}
\makeatother

\newlength{\thickarrayrulewidth}
\usepackage{booktabs}
\usepackage{balance}
\usepackage{multirow}
\usepackage{overpic}
\usepackage{cite}
\usepackage[font=small,labelfont=bf]{caption}
\usepackage{color}
\usepackage{pifont}
\makeatletter
\@namedef{ver@everyshi.sty}{}
\makeatother
\usepackage{tikz}
\usetikzlibrary{patterns}

\usepackage{colortbl}
\usepackage{pgfplots}
\pgfplotsset{compat=1.17}
\usepackage{tabularx}
\usepackage{arydshln}
\usepackage{amssymb}%
\usepackage{pifont}%

 \usepackage{amsmath}
\usepackage{MnSymbol}%
\usepackage{wasysym}%
\usepackage{enumitem}
\usepackage{wrapfig}
\usepackage{xspace}
\usepackage{tabu}
\usepackage[super]{nth}
\usepackage{scalefnt}
\usepackage{dsfont}
\usepackage{graphicx,calc}
\usepackage{subcaption}
\usepackage{algorithm}
\usepackage{algpseudocode}

\definecolor{darkgreen}{RGB}{0,153,51}
\definecolor{linkgreen}{RGB}{52,130,48}
\definecolor{LightCyan}{rgb}{0.87,0.92,0.96}
\definecolor{m_green}{RGB}{233, 254, 187}
\definecolor{m_orange}{RGB}{255, 212, 121}
\definecolor{m_red}{RGB}{255, 190, 188}
\definecolor{m_violet}{RGB}{215, 131, 255}
\definecolor{m_blue}{RGB}{186, 234, 255}
\definecolor{m_brown}{RGB}{255,212,120}
\definecolor{m_lightgreen}{RGB}{212,251,122}
\definecolor{notetext}{rgb}{0.7,0,0}

\definecolor{model_pink}{RGB}{235, 106, 164}
\definecolor{model_orange}{RGB}{250, 194, 122}
\definecolor{model_green}{RGB}{164, 210, 162}
\definecolor{model_gray}{RGB}{120, 120, 120}
\definecolor{model_yellow}{RGB}{251, 231, 171}
\definecolor{model_purple}{RGB}{190, 146, 211}

\usepackage{amssymb}%
\usepackage{pifont}%

\newcolumntype{Y}{>{\centering\arraybackslash}X}
\newcolumntype{Z}{>{\raggedleft\arraybackslash}X}

\usepackage{array}
\newcolumntype{P}[1]{>{\centering\arraybackslash}p{#1}}
\newcolumntype{M}[1]{>{\centering\arraybackslash}m{#1}}

\definecolor{darkblue}{RGB}{60, 82, 145}
\definecolor{kingblue}{RGB}{65, 105, 225}

\definecolor{background}{RGB}{226, 226, 226}
\definecolor{head}{RGB}{210, 78, 142}
\definecolor{rightArm}{RGB}{255, 176, 0}
\definecolor{leftArm}{RGB}{228, 162, 227}
\definecolor{rightForeArm}{RGB}{90, 64, 210}
\definecolor{leftForeArm}{RGB}{243, 232, 88}
\definecolor{rightHand}{RGB}{158, 143, 20}
\definecolor{leftHand}{RGB}{192, 100, 119}
\definecolor{torso}{RGB}{100, 143, 255}
\definecolor{hips}{RGB}{129, 103, 106}
\definecolor{rightUpLeg}{RGB}{243, 115, 68}
\definecolor{leftUpLeg}{RGB}{152, 200, 156}
\definecolor{rightLeg}{RGB}{149, 192, 228}
\definecolor{leftLeg}{RGB}{152, 78, 163}
\definecolor{rightFoot}{RGB}{129, 0, 50}
\definecolor{leftFoot}{RGB}{76, 134, 26}

\newlength\myheight
\newlength\mydepth
\settototalheight\myheight{Xygp}
\begin{document}

\begin{center}
\maketitle
\end{center}

\begin{abstract}
    Multilingual large language models (LLMs) have gained significant popularity for their ability to process and generate text across multiple languages. However, deploying these models in production can be inefficient when only a subset of the supported languages is of interest. There has been some research conducted on identifying whether machine translation models have language-specific or language-agnostic heads, however no research has been conducted for multilingual LLMs, to the best of our knowledge, that as we know are capable of performing diverse tasks beyond just translation. This paper explores whether multilingual LLMs have specialized language attention heads for each language, and investigates the possibility of removing language-specific heads for unwanted languages without degrading performance in the targeted languages. Our findings could inform more efficient deployment strategies for multilingual LLMs, enabling reduced model complexity while maintaining high accuracy for targeted languages.
\end{abstract}

\section{Introduction}
\label{sec:introduction}

The advent of large language models (LLMs) has dramatically transformed the landscape of natural language processing (NLP), enabling machines to comprehend and generate human-like text across diverse languages. This shift has led to remarkable advancements in machine translation, text summarization, and conversational AI, among other applications. Early iterations of LLMs predominantly focused on monolingual tasks, with English being the most common target language. However, as the demand for multilingual capabilities surged, particularly in a globally interconnected world, there was a strong push toward the development of models capable of handling multiple languages simultaneously. Multilingual LLMs were born from this demand, promising to bridge linguistic divides and offer high-quality language processing across a wide array of languages.

Multilingual LLMs, such as Cohere and other prominent models, have emerged as critical tools in addressing underrepresented languages, providing a more inclusive platform for natural language understanding. These models are trained on diverse datasets encompassing numerous languages, allowing them to generalize across linguistic barriers and perform tasks ranging from text classification to translation with impressive accuracy. However, despite their wide-reaching capabilities, deploying such models in production environments often presents inefficiencies. A significant challenge arises when only a subset of the languages these models support is of interest, as the models retain unnecessary complexity by maintaining attention heads for all languages, regardless of whether they are needed in specific use cases.

A central question that arises from this observation is whether multilingual LLMs have specialized language attention heads that are specific to certain languages. If these language-specific attention heads do exist, it raises the possibility of pruning or removing the attention heads associated with unwanted languages without compromising the model's performance in the languages that matter most. Such an approach could not only reduce the computational overhead and memory footprint of these models but also enhance their efficiency when deployed in production settings where only a limited number of languages are required. To date, while there has been research into language-specific or language-agnostic attention heads in machine translation models, similar explorations for multilingual LLMs remain sparse.

This paper seeks to address this gap in the research by investigating whether multilingual LLMs exhibit specialized attention heads for individual languages, or if they operate with more language-agnostic mechanisms that generalize across languages. Specifically, we aim to examine the attention heads of a popular multilingual LLM, analyzing their behavior to determine whether pruning unwanted language heads is a viable strategy for streamlining model deployment. By understanding the structure and function of these attention heads, we can provide insights into how multilingual LLMs process different languages and whether model simplification can be achieved without sacrificing performance in the languages of interest.

Our findings could have significant implications for the deployment of multilingual LLMs in real-world applications, particularly in settings where language coverage needs to be optimized for specific user bases or geographic regions. Moreover, this study contributes to the broader field of mechanistic interpretability in neural networks, offering a clearer view of how multilingual models handle the complexities of language and enabling more efficient, targeted use of these powerful tools in the future.

\begin{figure}
  \centering
  \includegraphics[width=16cm]{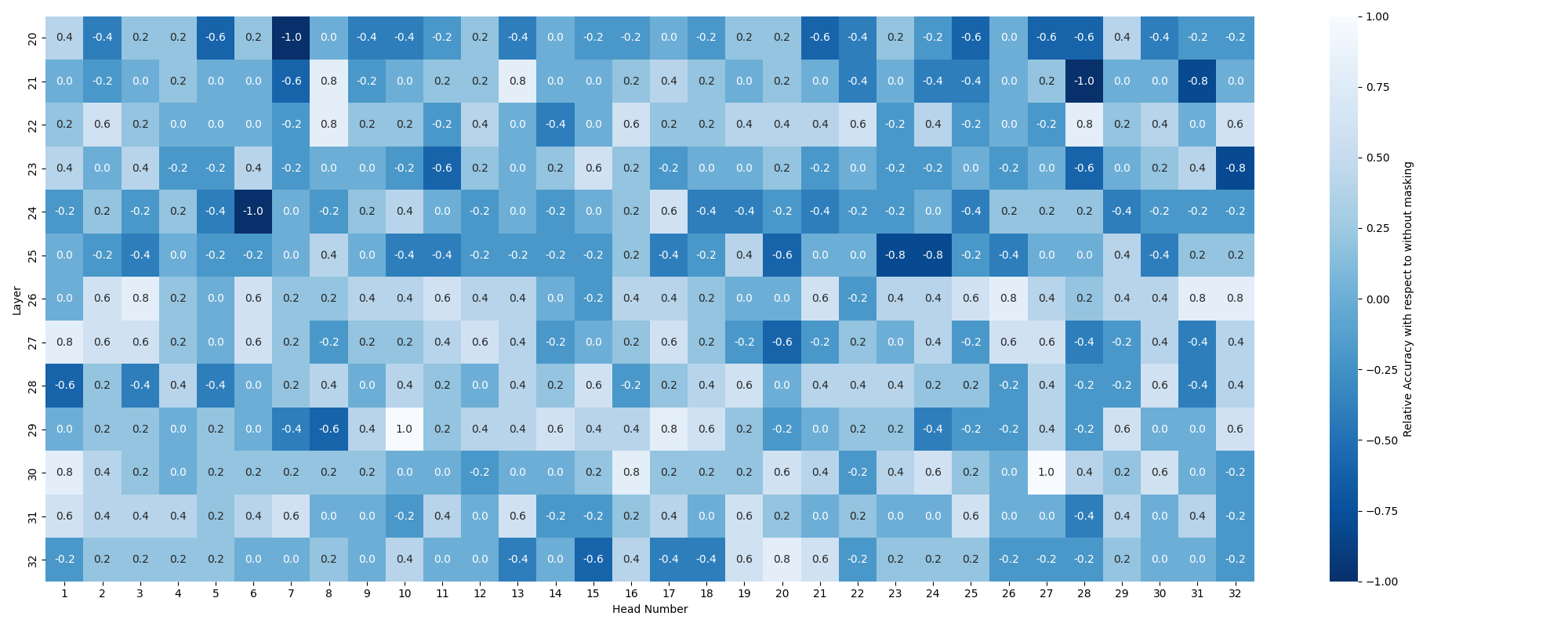}
  \includegraphics[width=16cm]{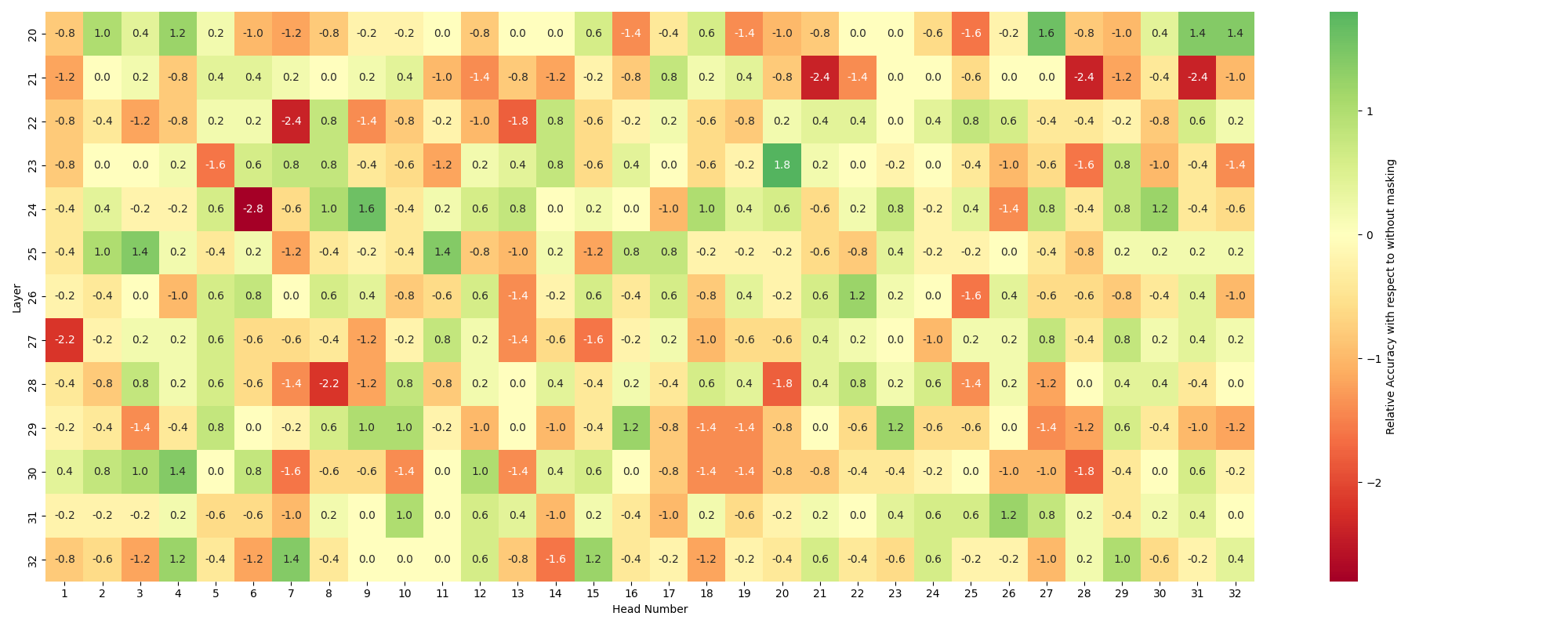}
  \caption{Heatmap of quantitative evaluation of Masking Heads Performance compared to No Masking. Top: English, Bottom: Hindi}
  \label{quantitative evaluation}
\end{figure}

\begin{figure}
  \centering
  \includegraphics[width=15cm]{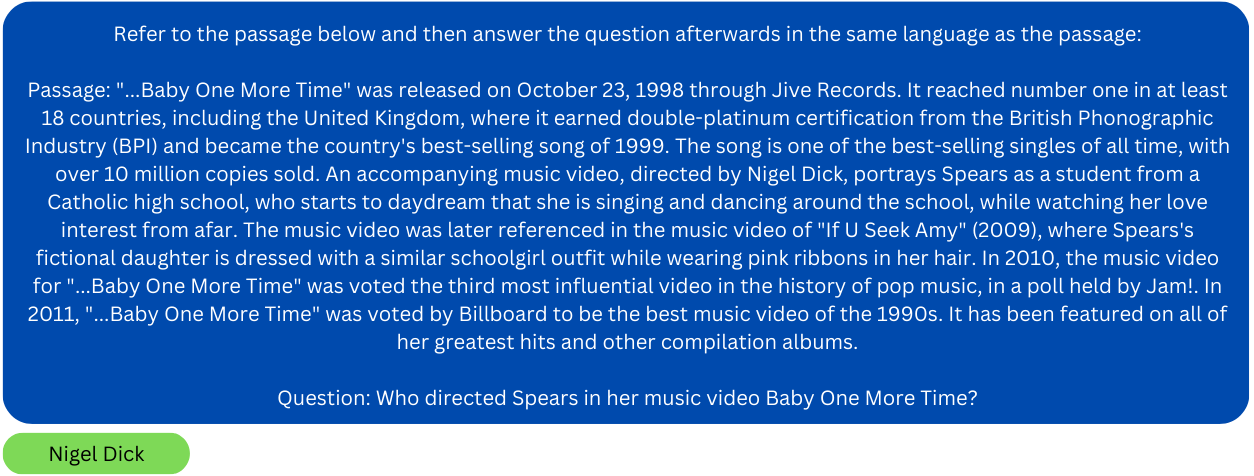}
  \includegraphics[width=15cm]{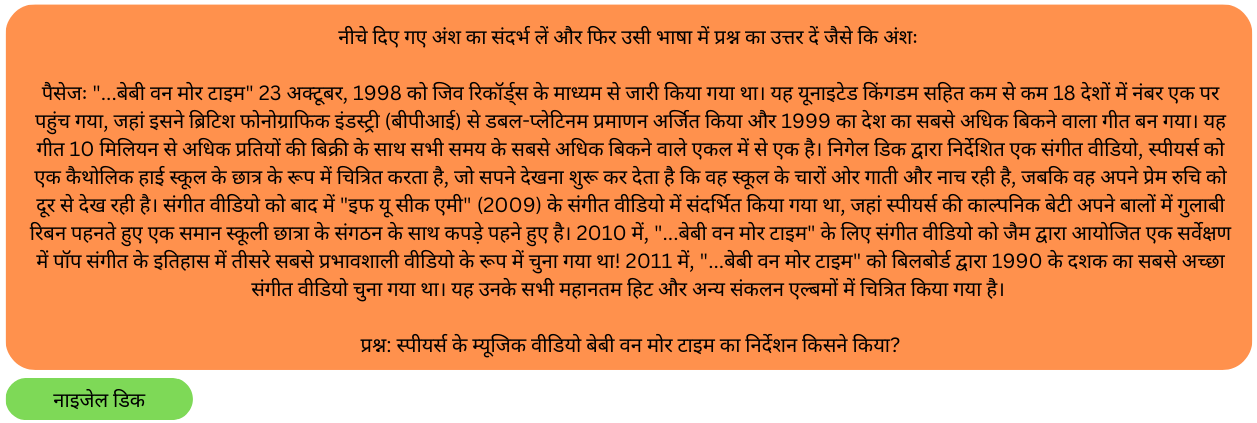}
  \caption{Aligned example. Top: English, Bottom Hindi}
  \label{aligned_examples}
\end{figure}

\section{Related work}
\label{sec:related_work}

\textbf{Mechanistic Interpretability.} Mechanistic interpretability aims to make neural network interpretable by breaking them down into several building blocks that can be well understood / and demystify the function of each. In this regard, a lot of research has been conducted to understand how language models work under the hood, rather treating them as a block box. Among those, some interpretable methods aren't scalable, like while others have proved to be fairy scalable. 

Anthropic team proposed a mathematical framework of a transformer\cite{elhage2021mathematical}, and showed that some attention heads (known as induction heads) have a tendency to pay attention and copy from similar tokens in the past, which might explain the in-context learning phenomenon in a transformer. 

Some other notable works include Logit Lens\cite{logitlens}, which shows how a language model refine its predictions gradually from early to final layers. OthelloGPT\cite{othelogpt} shows that a model trained on Othelo board game to predict a legal move is fully aware of the board state, also called world representation. The authors show that by finetuning a linear probe on the world representation, the model is able to predict the board state via a linear probe, indicating that the rest of the frozen network is well aware of the board state. Anthropic team pushed forward an interesting insight about the polysemantic nature of the neurons, suggesting that each neuron is activating for multiple kind of inputs, and thus isn't monosemantic. Breaking these polysemantic neurons into monosemantic ones is important, because this gives us more granular control of our model, as it allows us to drive the model output by let's say maximizing a hate neuron, or minimizing the neuron responsible for detecting bugs in the code, and thereby pushing model to give us output of the input code regardless of the bugs in it. Anthropic team achieved monosemanticity by training a sparse autoencoder\cite{monosemanticity, scaling_monosemanticity}. However, its application isn't limited to language models only. Recently, SAEs were applied on the early layers of Inception Network\cite{sae_cv} to identify features detected by them. As expected, the authors found out that early layers have learned basic features like edges and corners, which aligns with what we know about how computer vision models work; the early layers learn basic features like lines, edges, corners, while the later layers learn rather more complex features like eyes, ears depending on the problem they are trained for. 

RASP\cite{rasp} opensourced a programming language that simulates how transformer encoder process data, making it easier to understand their behavior by abstracting their computations into simple symbolic programs. Transformer Feed-Forward Layers Are Key-Value Memories\cite{geva2020transformer} argued that more than two-thirds parameters of a transformer-based language model are contained by neurons, yet little work is done on interpreting them. This served as a motivation for them to uncover the role of MLPs in a transformer-based language model. Authors suggested that neurons in a feed-forward layers operate as key-value pairs, where keys are activated on specific patterns in the input text like time, day, while values generate a distribution over possible next words or outputs based on the input patterns recognized by the keys.

\textbf{Masking/Knocking attention heads.} Michel et al.\cite{michel2019sixteen} found that only a small subset of attention heads significantly contribute to generating the output tokens. By pruning up to 50\% of these heads, they were able to maintain high accuracy comparable to the unpruned model. This result aligns with the findings of the lottery ticket hypothesis\cite{frankle2018lottery}, which proposes that overparameterized neural networks contain subnetworks—referred to as "winning tickets"—that can achieve performance similar to the full network when trained in isolation. Some neurons are initialized in a way that enables them to form these effective subnetworks, playing a crucial role in making predictions. The remaining neurons (or subnetworks) are redundant and can be pruned without significantly impacting performance. However, it is essential that the model is initially overparameterized to allow these winning subnetworks to emerge. Pruning must be performed after training; starting with a smaller subnetwork from the outset does not yield the same performance. Voita et al.\cite{voita-etal-2019-analyzing} showed that remaining attention heads start to assume responsibilities of multiple heads when other heads in the network are knocked out. 

\textbf{Language specific attention heads.} A cross-lingual task in natural language processing (NLP) refers to applying a model trained in one language (or a set of languages) to perform tasks such as text classification, translation, or information extraction in another language. The objective is for the model to leverage knowledge from the source languages and transfer it to target languages, often without requiring substantial training data in the target language. In this context, Shapley Head Pruning \cite{held-yang-2023-shapley} demonstrated that pruning language-specific attention heads can significantly reduce interference between languages, enabling better performance on languages the model was not explicitly trained on. Kim et al.\cite{kim2021multilingual} analyzed attention heads of a machine translation model, and the results reveal that, the most crucial attention heads are quite similar across different language pairs. Additionally, almost one-third of the less significant heads can be pruned with minimal impact on translation quality.

However it is not clear, how attention heads perform in a multilingual LLM. We analyze attention heads of the Cohere Model trained on 11 set of languages, and try to identify whether such models have language-specific or language-agnostic attention heads.


\vspace{-6px}

\section{Method}\label{sec:method}

\textbf{CohereModel} To conduct a research on a multilingual model, Cohere Model is a perfect choice. Cohere Model has 2 variants: Aya-101 and Aya-23. Aya-101\cite{aya101}, as evident from its name, is trained on 101 languages. However, its team realized that adding bulk of languages are hurting the model's performance. Therefore, in the Aya-23 model\cite{aya23}, they focused on depth instead of breadth of languages, and confined the model's scope to 23 languages only. Two of them are English and Hindi that we have carefully selected for this experiment. As per the numbers reported in Aya 23\cite{aya23}, English has the most speakers i.e. 500 million followed by Hindi i.e 350 million. It offers 8B and 35B params model. We have conducted experiments on the 4 bit quantized version of the 8B model. These models are trained on the Aya Dataset\cite{ayadataset}. This dataset was collected with the contributions of a huge family of researchers from all around the world who contributed to the dataset in their own languages. 

\textbf{Aya Dataset} We have evaluated the model on the "MLQA-en (T)" test split of the Aya Dataset\cite{ayadataset}, as it offers aligned examples in both languages (English and Hindi) to ensure a fair and balanced evaluation. Unaligned examples could raise concerns, as the English split may disproportionately focus on analytical or comprehension-based questions, while the Hindi split might emphasize different linguistic or cultural nuances, potentially skewing the assessment of cross-lingual model performance. Aligned example is shown in Figure \ref{aligned_examples}

\textbf{LLM as a Judge} GPT-3.5-Turbo is used a judge. Overall evaluation costed about \$150 with the following system prompt: 

\textit{You are a judge tasked with evaluating the accuracy of an LLM-generated response. You will be provided with a question from a passage, an LLM response answering the question, and the ground truth. Your job is to determine if the LLM response accurately conveys the essential meaning and key information from the ground truth, based on the question provided. }

\textit{
1. If the LLM response includes all key information from the ground truth (even if phrased differently or some words are implied based on the question), return (a score of) 1.}

\textit{
2. If the LLM response omits or misrepresents key information that changes the essential meaning, return (a score of) 0.}

\textit{**Important**: In some cases, information in the question itself (such as units, time references like "after," or context) may not need to be explicitly repeated in the LLM response if it is implied by the question.}

\textit{Your evaluation should focus on the meaning of the LLM response and whether it accurately answers the question based on the ground truth, without requiring the exact repetition of minor wording details. Please only return 1 or 0, without additional text.}
\section{Experiments}
\label{sec:experiments}
In order to maskout attention heads one by one, we have followed the similar mechanism as that of "Are Sixteen Heads Really Better than One?"\cite{michel2019sixteen}. The gate $G_h=\{0, 1\}$ can be controlled manually for each attention head $Att_h$, which allows to evaluate the model without that specific attention head contribution.
    \begin{equation}
        \MHAtt(x, q) = \sum_{h=0}^{N_h} G_h Att_h(x, q)
    \end{equation}

The model was evaluated rigorously over the last few layers. There is no special reason why only the last few layer, except for the reason to conclude the project early. The model was evaluated \textit{12 layers * 32 attention heads} times on 500 English examples and then again \textit{12 layers * 32 attention heads} times on 500 Hindi examples. 

We leveraged GPT-3.5-Turbo as a judge to ensure the LLM's response semantically contains the ground truth. It was used as a judge to strike a balance between per token cost of API and a reasonable accuracy. Given that we already provide the ground truth to the GPT, it doesn't need much creativity or knowledge of its own to give its decision whether LLM's response is correct.

The quantitative evaluation is shown in Figure \ref{quantitative evaluation}. Some qualitative results are discussed in the next section.

\section{Analyzing Attention Heads Behaviour}
\label{sec:experiments}
Language-specific are attention heads within a transformer model that focus on learning patterns and structures that are unique to a specific language. In other words, they are more tuned to the syntactic, semantic, or grammatical properties of a particular language. In contrast, language-agnostic attention heads refer to specific attention heads in a transformer model that function similarly across different languages. These attention heads are responsible for processing linguistic features that are common to multiple languages, such as sentence structure, syntax, or universal language patterns. Since they do not specialize in language-specific nuances, they can generalize well across various languages, allowing the model to perform well in multilingual tasks.

From the heatmap evaluation (Figure \ref{quantitative evaluation}), we can observe some \textit{language-specific attention heads} that are only making contribution (either positive or negative) in English language (also see Table \ref{tab:eng_lang_specific}), and there are some that are making contribution in just Hindi language (also see Table \ref{tab:hindi_lang_specific}). We can also observe \textit{language-agnostic attention heads} that are making contributions in both languages (also see Table \ref{tab:lang_agnostic}). In addition, there are some attention heads that don't affect the model's performance on either language (also see Table \ref{tab:misc_heads}), and there could be multiple explanations for that. Those could perhaps be language-specific attention heads that only play contribution for let's say Japanese, which we haven't evaluated our model on. Those could perhaps be language-agnostic attention heads, and only activate for specific patterns which our 500 aligned examples don't cover. Evaluating the model on a massive aligned examples may show their contribution. Lastly, they could be backup heads, that are only activated when others are knocked out, as we learn from Voita et al.\cite{voita-etal-2019-analyzing}. We call them \textit{miscellaneous heads}.

\section{Qualitative Results}

Some qualitative results are in Figure \ref{quantitative evaluation}. Following, we discuss the qualitative results.

\textbf{Example 1:} The following example shows how masking out some attention heads made the model make mistake, which otherwise didn't.

\textit{Refer to the passage below and then answer the question afterwards in the same language as the passage:}  

\textit{Passage: A person can be exposed to uranium (or its radioactive daughters, such as radon) by inhaling dust in air or by ingesting contaminated water and food. The amount of uranium in air is usually very small; however, people who work in factories that process phosphate fertilizers, live near government facilities that made or tested nuclear weapons, live or work near a modern battlefield where depleted uranium weapons have been used, or live or work near a coal-fired power plant, facilities that mine or process uranium ore, or enrich uranium for reactor fuel, may have increased exposure to uranium. Houses or structures that are over uranium deposits (either natural or man-made slag deposits) may have an increased incidence of exposure to radon gas. The Occupational Safety and Health Administration (OSHA) has set the permissible exposure limit for uranium exposure in the workplace as 0.25 mg/m3 over an 8-hour workday. The National Institute for Occupational Safety and Health (NIOSH) has set a recommended exposure limit (REL) of 0.2 mg/m3 over an 8-hour workday and a short-term limit of 0.6 mg/m3. At levels of 10 mg/m3, uranium is immediately dangerous to life and health.Most ingested uranium is excreted during digestion. Only 0.5\% is absorbed when insoluble forms of uranium, such as its oxide, are ingested, whereas absorption of the more soluble uranyl ion can be up to 5\%. However, soluble uranium compounds tend to quickly pass through the body, whereas insoluble uranium compounds, especially when inhaled by way of dust into the lungs, pose a more serious exposure hazard. After entering the bloodstream, the absorbed uranium tends to bioaccumulate and stay for many years in bone tissue because of uranium's affinity for phosphates. Uranium is not absorbed through the skin, and alpha particles released by uranium cannot penetrate the skin.}  

\textit{Question: Which type of substance causes greater risks when exposed to?}

\textit{Ground Truth: insoluble uranium compounds}

Table \ref{tab:llm_1_masked_0} shows the results of masking in response to the above question. This example explains how masking out some attention heads are resulting in decrease in accuracy.

\begin{table}[h]
\centering
\begin{tabular}{lcccccccccccccccccccccccccccccccc}
\toprule
Layer & Masked Head & LLM Response \\
\midrule
20 & 5 & soluble uranium compounds \\
20 & 6 & soluble uranium compounds \\
21 & 25 & soluble uranium compounds \\
22 & 3 & soluble uranium compounds \\
23 & 32 & soluble uranium compounds \\
24 & 11 & soluble uranium compounds \\
26 & 11 & soluble uranium compounds \\
32 & 18 & soluble uranium compounds \\
\midrule
No Masking & \ & insoluble uranium compounds \\
\bottomrule
\end{tabular}
\caption{Model making mistake upon masking}
\label{tab:llm_1_masked_0}
\end{table}

\textbf{Example 2:} The following example shows how masking out some attention heads made the model correct its mistake. We learn from 

\textit{Refer to the passage below and then answer the question afterwards in the same language as the passage:}

\textit{Passage: A similar struggle began in India when the Government of India Act 1919 failed to satisfy demand for independence. Concerns over communist and foreign plots following the Ghadar conspiracy ensured that war-time strictures were renewed by the Rowlatt Acts. This led to tension, particularly in the Punjab region, where repressive measures culminated in the Amritsar Massacre. In Britain public opinion was divided over the morality of the massacre, between those who saw it as having saved India from anarchy, and those who viewed it with revulsion. The subsequent Non-Co-Operation movement was called off in March 1922 following the Chauri Chaura incident, and discontent continued to simmer for the next 25 years.In 1922, Egypt, which had been declared a British protectorate at the outbreak of the First World War, was granted formal independence, though it continued to be a British client state until 1954. British troops remained stationed in Egypt until the signing of the Anglo-Egyptian Treaty in 1936, under which it was agreed that the troops would withdraw but continue to occupy and defend the Suez Canal zone. In return, Egypt was assisted in joining the League of Nations. Iraq, a British mandate since 1920, also gained membership of the League in its own right after achieving independence from Britain in 1932. In Palestine, Britain was presented with the problem of mediating between the Arabs and increasing numbers of Jews. The 1917 Balfour Declaration, which had been incorporated into the terms of the mandate, stated that a national home for the Jewish people would be established in Palestine, and Jewish immigration allowed up to a limit that would be determined by the mandatory power. This led to increasing conflict with the Arab population, who openly revolted in 1936. As the threat of war with Germany increased during the 1930s, Britain judged the support of Arabs as more important than the establishment of a Jewish homeland, and shifted to a pro-Arab stance, limiting Jewish immigration and in turn triggering...}

\textit{Question: Which treaty did Canada refuse to agree to in 1923?}

\textit{Ground Truth: the Treaty of Lausanne}

Table \ref{tab:llm_0_masked_1} shows the results of masking in response to the above question. This example explains how masking out some attention heads are resulting in increase in accuracy.
 
\begin{table}[h]
\centering
\begin{tabular}{lcccccccccccccccccccccccccccccccc}
\toprule
Layer & Masked Head & LLM Response \\
\midrule
20 & 27 & the Treaty of Lausanne \\
21 & 1 & the Treaty of Lausanne \\
22 & 31 & the Treaty of Lausanne \\
22 & 32 & the Treaty of Lausanne \\
24 & 15 & the Treaty of Lausanne \\
\midrule
No Masking & \ & the League of Nations \\
\bottomrule
\end{tabular}
\caption{Model correcting its mistake upon masking}
\label{tab:llm_0_masked_1}
\end{table}

\begin{table}[!htb]
    \begin{minipage}{.5\linewidth}

      \centering
        \begin{tabular}{ll}
        \toprule
            Layer & Head Number \\
            \midrule
            20 & 14 \\
            21 & 23 \\
            21 & 26 \\
            23 & 2 \\
            23 & 22 \\
            27 & 23 \\
            29 & 6 \\
            29 & 21 \\
            30 & 11 \\
            31 & 9 \\
            32 & 9 \\
            32 & 11 \\
            \bottomrule
        \end{tabular}
        \caption{Miscillineaous Heads}
        \label{tab:misc_heads}
    \end{minipage}%
    \begin{minipage}{.5\linewidth}
      \centering
        \begin{tabular}{ll}
            \toprule
            Layer & Head Number \\
            \midrule
            20 & 1 \\
            20 & 2 \\
            20 & 3 \\
            20 & 2 \\
            20 & 4 \\
            20 & 5 \\
            . & . \\
            . & . \\
            \bottomrule
        \end{tabular}
        \caption{Language Agnostic Attention Heads. Dots represent the list goes on}
        \label{tab:lang_agnostic}
    \end{minipage} 
\end{table}

\begin{table}[!htb]
    \begin{minipage}{.5\linewidth}

      \centering
        \begin{tabular}{ll}
        \toprule
            Layer & Head Number \\
            \midrule
            20     &      11 \\
            20     &      13 \\
            20     &      22 \\
            20     &      23 \\
            21     &       2 \\
            21     &       8 \\
            21     &      24 \\
            21     &      27 \\
            22     &      23 \\
            23     &       3 \\
            23     &      17 \\
            23     &      24 \\
            24     &      14 \\
            24     &      16 \\
            25     &      26 \\
            26     &       3 \\
            26     &       7 \\
            26     &      24 \\
            28     &      13 \\
            28     &      28 \\
            28     &      32 \\
            29     &      13 \\
            29     &      26 \\
            30     &       5 \\
            30     &      16 \\
            30     &      25 \\
            30     &      30 \\
            31     &      11 \\
            31     &      22 \\
            31     &      32 \\
            32     &      10 \\
            \bottomrule
        \end{tabular}
        \caption{English Language Specific Heads}
        \label{tab:eng_lang_specific}
    \end{minipage}%
    \begin{minipage}{.5\linewidth}
      \centering
        \begin{tabular}{ll}
            \toprule
            Layer & Head Number \\
            \midrule
            20 & 8 \\
            20 & 17 \\
            20 & 26 \\
            21 & 1 \\
            21 & 3 \\
            21 & 5 \\
            21 & 6 \\
            21 & 10 \\
            21 & 14 \\
            21 & 15 \\
            . & . \\
            . & . \\
            \bottomrule
        \end{tabular}
        \caption{Hindi Language Specific Heads. Dots represent the list goes on}
        \label{tab:hindi_lang_specific}
    \end{minipage} 
\end{table}
\section{Conclusion}
\label{sec:discussion}
We evaluate the multilingual LLM's performance across 2 difference languages, and identify their behaviour after masking out attention heads iteratively. We show that there are 4 types of attention heads: English language-specific, Hindi language-specific, language-agnostic and Miscellaneous. We present rigorous qualitative and quantitative results to support the findings.

\clearpage
\balance

{\small
\bibliographystyle{plainnat}
\bibliography{egbib}

\newpage
\clearpage




}

\end{document}